\pdfoutput=1

\documentclass[11pt]{article}

\usepackage{EMNLP2023}

\usepackage{times}
\usepackage{latexsym}

\usepackage[T1]{fontenc}

\usepackage[utf8]{inputenc}

\usepackage{microtype}

\usepackage{inconsolata}
\usepackage{comment}
\usepackage{times}
\usepackage{pifont}
\usepackage[normalem]{ulem}
\usepackage{latexsym}
\usepackage{amsmath,amsfonts,amssymb}
\usepackage{subcaption}
\usepackage[T1]{fontenc}

\usepackage[utf8]{inputenc}

\usepackage{microtype}
\usepackage{paralist}

\usepackage{enumitem}

\usepackage{xspace}
\usepackage{hyperref}
\usepackage{url}
\usepackage{wrapfig}
\usepackage{graphicx}
\usepackage{caption}
\usepackage{subcaption}
\usepackage{multirow}
\usepackage{booktabs}
\usepackage{algorithm}
\usepackage{algpseudocode}
\usepackage{amsmath}
\usepackage{amsfonts}
\usepackage{comment}
\usepackage{makecell}
\usepackage{xcolor}

\usepackage{arydshln}
\usepackage{amsthm}

\newcommand{\ifc}{\textit{in-file context}\xspace}
\newcommand{\cfc}{\textit{cross-file context}\xspace}

\newcommand{\cmark}{\ding{51}}%
\newcommand{\xmark}{\ding{55}}%
\newcommand{\mycmark}{\color{teal} \cmark \color{black}}
\newcommand{\myxmark}{\color{red} \xmark \color{black}}

\newcommand{\eg}{\hbox{\textit{e.g.,}}\xspace}
\newcommand{\ie}{\hbox{\textit{i.e.,}}\xspace}
\newcommand{\wrt}{\hbox{\textit{w.r.t.}}\xspace}

\newcommand{\tool}{\textsc{CoCoMIC}\xspace}
\newcommand{\ccfinder}{\textsc{CCFinder}\xspace}

\usepackage{fnpct}

\usepackage{listings}
\usepackage{adjustbox}
\usepackage[most]{tcolorbox}

\definecolor{dark-gray}{gray}{0.85}
\definecolor{light-gray}{gray}{0.95}
\definecolor{mygreen}{rgb}{0,0.4,0}
\definecolor{mygray}{rgb}{0.5,0.5,0.5}
\definecolor{mymauve}{rgb}{0.58,0,0.82}
\definecolor{myred}{rgb}{0.82, 0.1, 0.26}

\lstdefinestyle{CustomPy}{
    escapeinside={(*@}{@*)},
    belowcaptionskip=1\baselineskip,
    xleftmargin=1pt,
    xrightmargin=1pt,
    language=Python,
    numbersep=5pt,
    tabsize=4,
    showstringspaces=false,
    basicstyle=\footnotesize\ttfamily,
    keywordstyle=\bf\color{mygreen},
    commentstyle=\color{purple},
    stringstyle=\color{red},
    identifierstyle=\color{black},
    numberstyle=\tiny\color{mygray},
    emph={CURSOR_POSITION},
    emphstyle={\bf\color{myred}},
    emph=[2]{and, in,},
    emphstyle=[2]{\bf\color{violet}},
    emph=[3]{sortedCount, sorted_count},
    emphstyle=[3]{\bf\color{blue}},
    numbers=left,
    stepnumber=1,
    breaklines=true,
    backgroundcolor=\color{white},
    literate={\ \ }{{\ }}1,
}
\definecolor{codegreen}{rgb}{0,0.6,0}
\definecolor{codegray}{rgb}{0.5,0.5,0.5}
\definecolor{codepurple}{rgb}{0.58,0,0.82}
\definecolor{backcolour}{rgb}{0.95,0.95,0.92}
\definecolor{green}{HTML}{268B07}
\definecolor{blue}{HTML}{4077ab}
\definecolor{red}{HTML}{CC8E7F}
\definecolor{magenta}{HTML}{A748C3}
\definecolor{redorange}{HTML}{F46A4E}

\lstdefinestyle{mystyle}{
    backgroundcolor=\color{backcolour},   
    commentstyle=\color{codegreen},
    keywordstyle=\color{magenta},
    numberstyle=\tiny\color{codegray},
    stringstyle=\color{codepurple},
    basicstyle=\scriptsize\ttfamily,
    ndkeywordstyle=\color{darkgray}\bfseries,
    identifierstyle=\color{black},
    breakatwhitespace=false,         
    breaklines=true,                 
    captionpos=b,                    
    keepspaces=true,                 
    numbers=left,                    
    numbersep=5pt,                  
    showspaces=false,                
    showstringspaces=false,
    showtabs=false,                  
    tabsize=2,
}
\makeatletter
\let\old@lstKV@SwitchCases\lstKV@SwitchCases
\def\lstKV@SwitchCases#1#2#3{}
\makeatother
\usepackage{lstlinebgrd}
\makeatletter
\let\lstKV@SwitchCases\old@lstKV@SwitchCases

\lst@Key{numbers}{none}{%
    \def\lst@PlaceNumber{\lst@linebgrd}%
    \lstKV@SwitchCases{#1}%
    {none:\\%
     left:\def\lst@PlaceNumber{\llap{\normalfont
                \lst@numberstyle{\thelstnumber}\kern\lst@numbersep}\lst@linebgrd}\\%
     right:\def\lst@PlaceNumber{\rlap{\normalfont
                \kern\linewidth \kern\lst@numbersep
                \lst@numberstyle{\thelstnumber}}\lst@linebgrd}%
    }{\PackageError{Listings}{Numbers #1 unknown}\@ehc}}
\makeatother

\newcount\myloopcounter

\newcommand{\repeatit}[2][10]{%
  \myloopcounter0%
  \loop\ifnum\myloopcounter < #1 %
  #2%
  \advance\myloopcounter by 1 %
  \repeat %
}

\usepackage{xfp}
\usepackage{anyfontsize}

\makeatletter
{\small %
\xdef\f@size@small{\f@size}
\xdef\f@baselineskip@small{\f@baselineskip}
\normalsize %
\xdef\f@size@normalsize{\f@size}
\xdef\f@baselineskip@normalsize{\f@baselineskip}
}
\newcommand{\smalltonormalsize}{%
  \fontsize
    {\fpeval{(\f@size@small+\f@size@normalsize)/2}}
    {\fpeval{(\f@baselineskip@small+\f@baselineskip@normalsize)/2}}%
  \selectfont
}
\makeatother

\makeatletter
\renewcommand{\ALG@beginalgorithmic}{\smalltonormalsize}
\makeatother

\usepackage{titlesec}
\titlespacing{\paragraph}{%
  0pt}{%
  0.2\baselineskip}{%
  1em}%

\title{\tool: \uwave{Co}de \uwave{Co}mpletion By Jointly \uwave{M}odeling \\ \uwave{I}n-file and \uwave{C}ross-file Context}

\author{Yangruibo Ding$^{1,*,\dagger}$ \quad Zijian Wang$^{2,*,\ddagger}$ \quad Wasi Uddin Ahmad$^{2,*}$  \\  \textbf{Murali Krishna Ramanathan}$^2$ \enspace  \textbf{Ramesh Nallapati}$^2$ \enspace\textbf{Parminder Bhatia}$^2$  \\ \enspace \textbf{Dan Roth}$^2$  \enspace \textbf{Bing Xiang}$^2$ \\
         $^{1}$Columbia University \quad $^{2}$AWS AI Labs  \\
\texttt{yrbding@cs.columbia.edu} \quad \texttt{\{zijwan,wuahmad\}@amazon.com}\\
\texttt{\{mkraman,rnallapa,parmib,drot,bxiang\}@amazon.com}}

\begin{document}
\maketitle

\begin{abstract}

While pre-trained language models (LM) for code have achieved great success in code completion, they generate code conditioned only on the contents within the file, \ie \ifc, but ignore the rich semantics in other files within the same project, \ie \cfc, a critical source of information that is especially useful in modern modular software development. Such overlooking constrains code language models’ capacity in code completion, leading to unexpected behaviors such as generating hallucinated class member functions or function calls with unexpected arguments. In this work, we develop a cross-file context finder, \ccfinder, that locates and retrieves the most relevant cross-file context. We propose \tool, a framework that incorporates cross-file context to learn the in-file and cross-file context jointly on top of pretrained code LMs. 
\tool successfully improves the existing code LM with a 33.94\% relative increase in exact match and a 28.69\% relative increase in identifier matching for code completion when the cross-file context is provided. %

\end{abstract}

\section{Introduction}

In recent years, language models for source code like Codex \cite{chen2021evaluating} and CodeGen \cite{nijkamp2022codegen} have shown promising performance in code completion tasks and have great potential to improve developer productivity \cite{barke2022grounded}. These code LMs are typically trained with causal language modeling loss and complete the code conditioning on the previous code tokens in the same file, which we refer to as \ifc.

Modular programming~\cite{parnas1972module,parnas1985modular,sullivan2001modularity} is a software design strategy that divides the complex software functionality into several independent, interchangeable components (\eg files, classes, and functions), such that each component implements only one aspect of the desired functionality and consequently becomes easily reusable and testable. It has already been a well-adapted paradigm in modern software development and maintenance. Developing under the modular programming paradigm requires knowledge from the current file and the whole project, to which we refer as \cfc.
As shown in Figure \ref{fig:motivation_example}, the \cfc is critical for code completion: the CodeGen Python model~\cite{nijkamp2022codegen} with 2 billion parameters fails to generate the correct code since it only considers \ifc and lacks visibility to various crucial references for code completion, \eg member functions of imported classes and arguments of imported functions. 

\begin{figure}[!tp]
    \centering
    \includegraphics[width=\columnwidth]{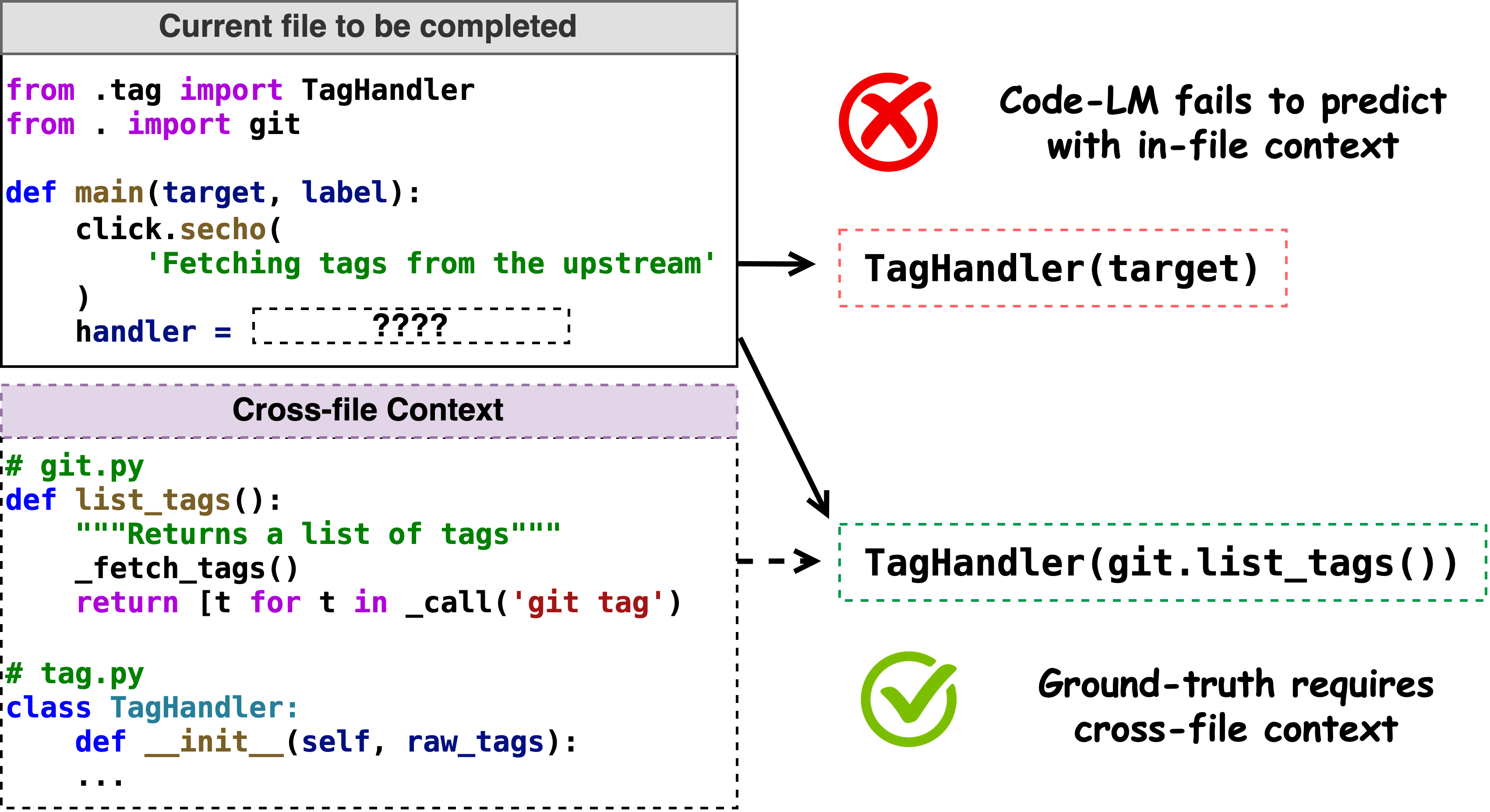}
    \caption{CodeGen-2B-mono fails to complete a Python program correctly as \ifc does not provide sufficient information. The model needs to know \texttt{TagHandler} takes an argument \texttt{raw\_tags}, which could be obtained through the function \texttt{list\_tags} of \texttt{git}. Generating the correct code requires the presence of class and function definitions as part of the context, which cannot be derived from the current file alone.}
    \label{fig:motivation_example}
    \vspace{-2mm}
\end{figure}

In this work, we argue that code LMs should generate code conditioned jointly on \ifc and \cfc.
However, there are challenges in developing such models. First, the project defines its individual and complex hierarchy and could be of varied sizes. Thus, given a piece of code, it is critical yet challenging to identify the most relevant and useful cross-file context. 
Second, we must carefully design a framework for aggregating the information from the in-file and cross-file context. Na\"ively concatenating code from in-file and cross-file context is not feasible for three reasons. (1) They represent distinct types of contextual information, as the former presents the local dependencies and human intentions (\eg code comments) for code completion, while the latter compensates for the project-level dependencies that do not exist in the surrounding lines. Thus, the model should not always treat them equally. (2) Unlike third-party packages which are mostly available in the pre-training dataset of LLMs, the same project context are likely to be private to the model, i.e., the model didn't see it during pre-training. This makes code completion that requires same project context very difficult if without the right context at inference time. (3) The model's input length is limited, so concatenating all context as input would exceed its context length capacity. 

To address the aforementioned challenges, we build  \ccfinder, a cross-file context finder that effectively retrieves the most relevant cross-file context given a code snippet to be completed. Furthermore, we propose \tool, a novel framework that jointly learns in-file and cross-file context to improve code completion.

\paragraph{Cross-file Context Finder} We design and implement \ccfinder, a static code analysis tool, to retrieve the most relevant cross-file context for code completion. \ccfinder parses the project hierarchy and code components to extract project information. \ccfinder further builds a project context graph to represent the details of each component (\ie entity) and the interactions among them (\ie relation). When an incomplete program requests completion, the tool will first analyze its \texttt{import} statements and pinpoint the related entities from the built context graph. Then, the tool will retrieve the neighbors of the pinpointed entities from the graph as the cross-file context of the current file. 

\paragraph{Jointly Modeling In-file and Cross-file Context} 
We propose \tool, a novel framework built on top of existing code LMs with joint attention to in-file and retrieved cross-file context. We realize this in two steps: 
First, the model will compress cross-file context and build its representations. 
Second, when generating code completion, the model will attend to both the compressed cross-file context and the concrete in-file context. 

We evaluate the effectiveness of \ccfinder and \tool on a code completion dataset we built from the Python Package Index (PyPI), a repository of open-source Python projects. We show that \ccfinder is able to retrieve 27.07\% more relevant context for code completion than in-file context. Experiments show that \tool with access to relevant cross-file context improves the backbone pretrained code LM, CodeGen \cite{nijkamp2022codegen}, by 33.94\% in exact match and 28.69\% identifier matches relatively.

\noindent Our main contributions are: 
\begin{enumerate}[itemsep=0pt,topsep=0pt]
    \item We present \tool, a novel framework built on top of code LMs that jointly learns in-file and cross-file context to enhance code completion. To power \tool, we develop \ccfinder, an effective tool at harvesting cross-file context, a critical yet overlooked resource for code completion in the era of modern software development.
    \item We show that \tool with cross-file context from \ccfinder significantly outperforms baselines by up to $+33.94\%$ in exact match. We additionally conduct extensive ablation studies to show the contribution of different components.
    \item We release a diverse and high-quality dataset on statement-level code completion that tests model's ability on making use of cross-file context to facilitate further research\footnote{\url{https://github.com/amazon-science/cocomic}}. 
\end{enumerate}

\section{Preliminaries}
\label{subsec:preliminaries}
For the convenience of discussion, we define concepts that will be used throughout the paper.

\begin{figure*}[ht]
    \centering
    \includegraphics[width=0.9\textwidth]{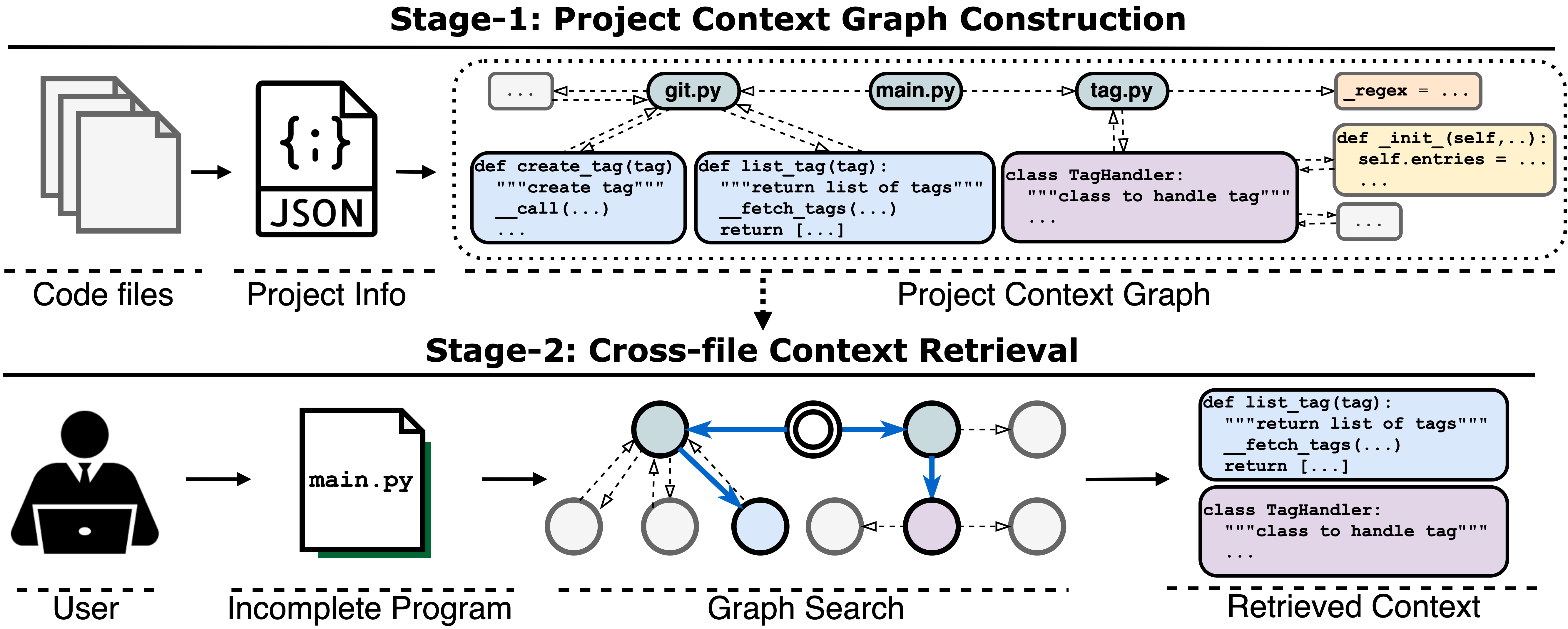}
    \caption{Overview of \ccfinder. First, \ccfinder builds the project context graph, including the bird's-eye view of the whole project and the code details of each module. Then, given the incomplete program, \ccfinder retrieves a set of the most relevant project entities as cross-file context from the graph.
    }
    \vspace{-3pt}
    \label{fig:context_constructions}
\end{figure*}

\paragraph{Project Entities} Project entities are code components that constitute the skeleton of software projects; developers frequently import and reuse these entities as cross-file context. We focus on four types of entities: \emph{file, function, class, and global variable}. In particular, \emph{file} contains the file name and file docstring;  \emph{class} contains the class signature, docstring, and attributes; \emph{function} contains function signature,  docstring, and body; \emph{global variable} contains the variable name and its value.

\paragraph{Entity Relations} Entity relations represent the interactions among project entities. We consider two categories of relations: \emph{intra-file} and \emph{inter-file}. 
Intra-file relations describe the in-file code hierarchies pre-defined by the programming language grammar. 
For example, a class is at the first level of the hierarchy while its member functions are at the second level.
Inter-file relations define the file-to-file dependencies. Under each category, we further define several types of relations (Appendix~\ref{subsec:appendix_edge_types}).

\paragraph{Locale} We define \emph{locale} as the entity's relative code location within the software project. For example, the locale of class entities is defined as \texttt{file\_name.class\_name}. The locale is assigned a unique name according to the specific location of a project entity, so we maintain the one-to-one mapping between each entity and its locale. The locale benefits \tool in two ways: (1) when we construct cross-file context, the locale efficiently maps the relative path of a code snippet to its project entity \ccfinder builds (\S\ref{subsec:node_retrieval}), and (2) it indicates hierarchical relations among project entities and helps model with code completion (\S\ref{subsec:impacts_cross_relation}).

\paragraph{In-file \& Cross-file Context} For an incomplete source file $\mathcal{S}$, we define two types of context: \emph{in-file} and \emph{cross-file}. In-file context represents code snippets included in the current file, \ie code tokens before the predicting position. Cross-file context $\mathcal{C}$ represents the relevant code information (\eg classes, functions) from the same project that is out of but imported by the current file. Concretely, \cfc refers to a collection of relevant project entities that might assist with the missing code prediction but are not in $\mathcal{S}$.

\section{Cross-file Context Finder: \ccfinder}
\label{sec:context_construction}

Software projects typically have complex structures \cite{parnas1985modular} representing the dependencies among distinct code components.  
To retrieve the most relevant code snippets given a code, we need a tool with two main characteristics. First, the tool should be able to navigate the project structure to identify the file and module dependencies. Second, the tool can zoom into the dependencies and extract detailed code components.
Off-the-shelf tools do not meet the requisites. For example, module dependency analysis tools\footnote{\small\url{https://github.com/google/importlab}}\footnote{\small\url{https://github.com/thebjorn/pydeps}} can only provide the module interactions while missing the hierarchical details inside each module and cannot directly output the concrete code. Therefore, we develop a new tool, \ccfinder, to aggregate cross-file context.

\ccfinder's overall workflow is shown in Figure~\ref{fig:context_constructions}. It has two main steps: (1) Analyze the program dependencies to build a bird's-eye view of the whole project and parse the source code to extract code details of each module. With these, \ccfinder builds the \emph{project context graph}: graph nodes represent code components that constitute the project's backbone, and edges indicate the relations among components. (2) Given an incomplete program, the tool retrieves the most relevant cross-file context from the built graph. In this work, we focus on 
Python as the proof-of-concept to showcase our main arguments. However,  \ccfinder's conceptual design is extensible to other languages. %

\subsection{Project Context Graph}
\label{subsec:context_graph}

\ccfinder parses the project structure and corresponding source files to identify the project entities and entity relations. Then, \ccfinder uses entities and entity relations to build graph nodes and directed edges, respectively. The context graph is built top-down. First, we create a root node for the project and connect it with all file nodes. Second, each file node will build its own sub-graph, wrapping code components within the file, and also build connections with other files that it depends on, \ie it imports code from these files. Third, nodes will link to others within the file-level sub-graph based on the dependencies or scope. For example, a class node will have edges to its member functions. 

Formally, \ccfinder builds the multi-relational, directed context graph $\mathcal{G} = (\mathcal{V}, \mathcal{E})$ for the project, where $\mathcal{V}$ is the set of nodes representing code components, and $\mathcal{E}\subseteq\mathcal{V}\times\mathcal{R}\times \mathcal{V}$ is the set of edges that indicate the interactions among code components, where $\mathcal{R}$ is the set of edge types (Appendix~\ref{subsec:appendix_edge_types}).

Note that the project context graph generated by \ccfinder differs from the traditional program dependence graph~\cite{ferrante1987pdg} and code property graph~\cite{yamaguchi2014cpg}, which are built to estimate and analyze the program execution behaviors statically. These graphs focus on data flows and control flows, while our graph represents the dependencies of different modules within the project. \ccfinder-generated graph is also different from the code knowledge graph~\cite{abdelaziz2021graph4code,abdelaziz2022blanca} as the latter combines API usage knowledge such as third-party documentation and StackOverflow questions and answers.

\subsection{Cross-file Context Retrieval}
\label{subsec:node_retrieval}

The project context graph represents the project hierarchies and interactions among code components, so the closer a graph neighbor to a specific node, the more relevant that neighbor is. For example, if the input code imports a class, the most useful information regarding this class, such as its member functions and the global variables it depends on, should be only 1 or 2 hops away. Thus, given an input code, we retrieve a set of relevant nodes from the context graph as the cross-file context.

\setlength{\textfloatsep}{5mm}
\begin{algorithm}[t]
\caption{Retrieve Cross-file context}
\label{alg:context_retrieval}
$\mathcal{F}$: Incomplete code file\\
$\mathcal{G}$: Project context graph\\
$\mathcal{P}$: Parsed project information\\
$k$: Maximum depth of graph search
\begin{algorithmic}[1]

\State $ctx\_nodes \gets $\O
\State $\mathcal{I} \gets GetLocalImportStmt(\mathcal{F})$
\For {$stmt \in \mathcal{I}$}
\State$root\gets LocateNode(\mathcal{G}, stmt)$
\State$\mathcal{N} \gets DepthFirstSearch(\mathcal{G}, root, k)$
\For{$n \in \mathcal{N}$}
\If{$n \notin ctx\_nodes$}
    \State $ctx\_nodes.add(n)$
\EndIf
\EndFor
\State$ReorderNode(ctx\_nodes, \mathcal{P})$
\EndFor\\
\Return{$ctx\_nodes$}
\end{algorithmic}
\end{algorithm}

The workflow is presented in Algorithm~\ref{alg:context_retrieval}. First, we extract the \texttt{import} statements from the incomplete code file ($\mathcal{F}$) that only imports code snippets within the same project ($GetLocalImportStmt$). We iteratively use each \texttt{import} to identify and locate corresponding nodes in the project context graph ($LocateNode$). With the direct mapping between the code snippets and their locales (\S\ref{subsec:preliminaries}), we can locate the node given its relative path. We use such a node as the root node and retrieve its neighbors within $k$ hops using the depth-first graph search ($DepthFirstSearch$). $k$ is a configurable hyper-parameter in which increasing $k$ will retrieve a broader context. 
We set $k = 2$ for all the experiments in this work 
 and empiricially justify the choice in \S\ref{subsec:ccfinder_effectiveness}.
Finally, once we collect the $k$-hop neighbors for all \texttt{import} statements, we re-order the nodes ($ReorderNode$), ensuring the nodes from the same source file follow the original code order, to maintain the naturalness~\cite{hindle-2012-naturalness} of human-written code.

\begin{figure*}
    \centering
    \includegraphics[width=0.925\textwidth]{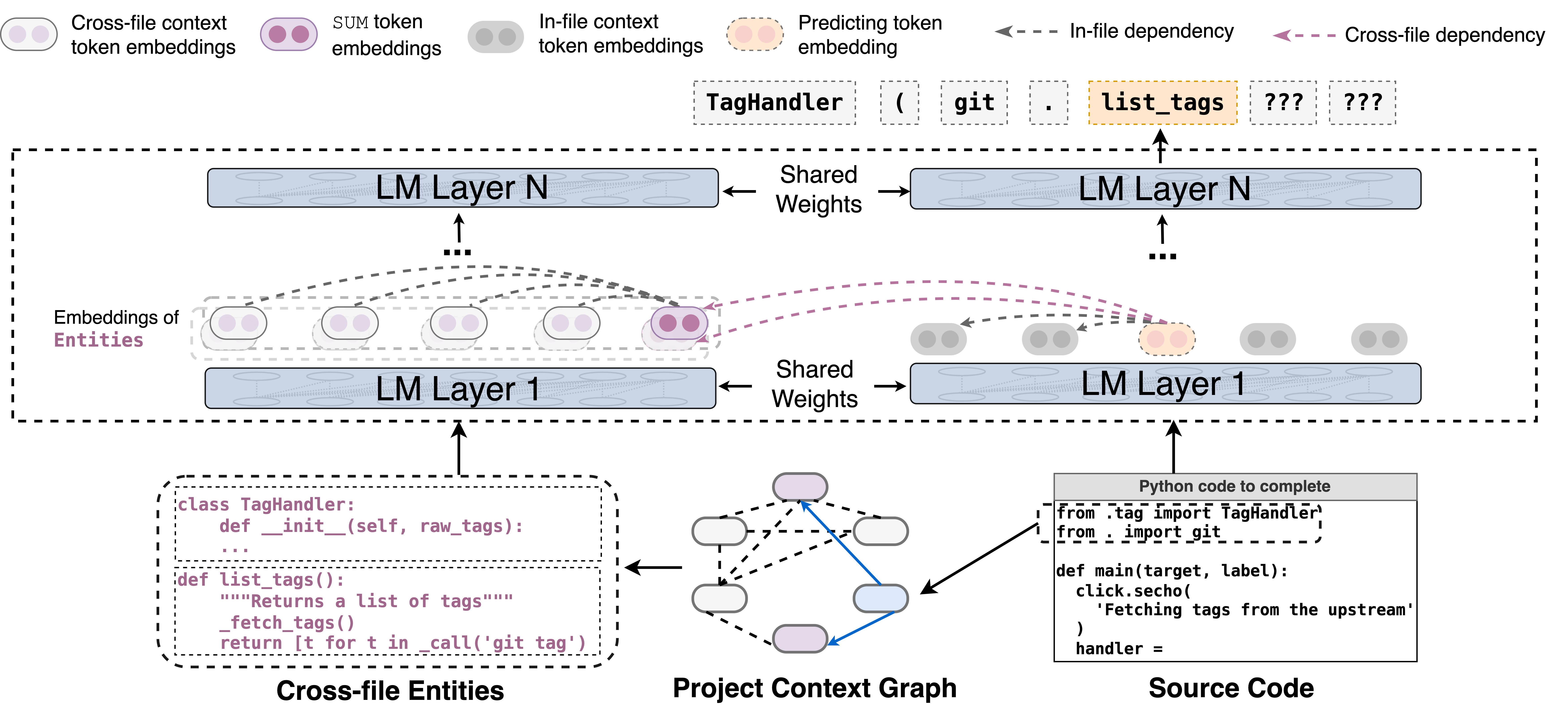}
    \caption{The \tool framework. \textbf{Bottom}: Given incomplete code, \tool leverages \ccfinder to identify the corresponding entities in the project context graph (\S\ref{subsec:context_graph}) and retrieve their k-hop neighbors as cross-file entities (\S\ref{subsec:node_retrieval}). \textbf{Up}: 
    \tool first generates representations for cross-file entities using the appended \texttt{[SUM]} token (\S\ref{subsec:encode_cross_file_cxt}). Then it completes the current code by jointly attending to in-file and cross-file context (\S\ref{subsec:model_infile_crossfile_cxt}).
    \vspace{-4pt}}
    \label{fig:model_arch}
\end{figure*}

\section{Proposed Framework: \tool}
\label{sec:cocomic}

A high-level overview of the \tool framework is presented in Figure~\ref{fig:model_arch}. 
\tool uses an autoregressive LM to encode (1) in-file code snippet and (2) retrieved cross-file context, and predicts the next code token conditioning on both.

\subsection{Input Representation}
\label{subsec:model_input}
As shown in Figure~\ref{fig:model_arch}, the model input includes two parts: source code sample $\mathcal{S}$ and its cross-file context $\mathcal{C}$. Specifically, the source code sample $\mathcal{S}$ consists of a sequence of tokens $x_1, ..., x_T$, where $x_t$ is a code token and $T$ is the length of $\mathcal{S}$; the cross-file context, as introduced in \S\ref{sec:context_construction}, is a list of entities, $\mathcal{C} = (c_1, ..., c_n)$, retrieved from the project context graph. Each entity, $c_i$, is a short piece of code sequence describing the details of that entity, \ie $c_i = (locale_i, w_i^1, ..., w_i^m, \texttt{[SUM]})$, where $w_i^j$ is a code token within the entity, $locale_i$ is the locale~(\S\ref{subsec:preliminaries}) of $c_i$, and \texttt{[SUM]} is a special token.

\paragraph{Representing Entity Relations with Locales} As introduced in \S\ref{subsec:preliminaries}, each project entity is paired with a locale that indicates its hierarchical relationship. We explore the benefits of prepending locales to provide entities with such relational hints (\S\ref{subsec:impacts_cross_relation}). Specifically, for each cross-file entity, we prepend its locale to its code text as a comment, followed by a new line character: for the example in Figure~\ref{fig:model_arch}, the retrieved entity \texttt{def list\_tags()} will be prepended with \texttt{\#git.list\_tags\char`\\n}. 

\paragraph{Better Entity Representation with \texttt{[SUM]}} We append a special token \texttt{[SUM]} to entity descriptions. We expect \texttt{[SUM]} token to learn the summarization of the entity since the causal attention~\citep{radford2019gpt2,brown2020gpt3} allows it to attend to all the previous tokens describing the entity. 
When completing code, the model will attend to the representations of the \texttt{[SUM]} tokens for each cross-file entity.
We compare it with mean pooling in \S\ref{subsec:sum} and show that \texttt{[SUM]} works better.

\subsection{Encoding Cross-file Context}
\label{subsec:encode_cross_file_cxt}

The computational cost of Transformers increases exponentially \wrt the input length, so it is impractical to prepend all the retrieved entities as plain text, as they typically contain thousands of tokens (Appendix \ref{appendix:data-prep}). Also, only a few keywords in an entity (e.g., identifiers) play an important role in assisting code completion.  Thus, \tool encodes each 
entity into a single token to balance the space limitation and the information need. 

\vspace{-10pt} %
\begingroup\small\begin{gather*}
    h_{c_i} = f_\theta(c_i) \in \mathbb{R}^{d_h}, H_\mathcal{C} = (h_{c_1}, ..., h_{c_n}) \in  \mathbb{R}^{n \times d_h}
\end{gather*}\endgroup
\vspace{-15pt}

Specifically, for each entity $c_i$, the model $f_\theta$ will encode its code sequence into one representation $h_{c_i}\in\mathbb{R}^{d_h}$, where $d_h$ is the hidden dimension. Then, \tool takes the hidden state of the last token, \texttt{[SUM]}, as the entity representation. Finally, the model will output a list of entity embeddings, $H_\mathcal{C}$, 
representing the retrieved cross-file context.

\subsection{Modeling In-file and Cross-file Context for Code Completion}
\label{subsec:model_infile_crossfile_cxt}
After getting representations of cross-file context, \tool continues to encode the in-file context and train the model to learn both context jointly. 

\paragraph{In-file Context} \tool utilizes the causal language model setting to support the code completion task, where each token will consider its former texts as in-file context. Specifically, the in-file context of source code $\mathcal{S}$, at time step $t$, will be $s_t = (x_1, ..., x_{t-1})$. We pass these tokens through the model and get the embeddings of each token to construct the representation of the in-file context.

\vspace{-15pt} %
\begingroup\small\begin{gather*}
    H_\mathcal{S}(t) = f_\theta(s_t) = f_\theta(x_1, ..., x_{t-1}) \in \mathbb{R}^{(t-1) \times d_h}
\end{gather*}\endgroup

\paragraph{Joint attention to In-file and Cross-file Context} Different layers of a Transformer model have been shown to capture different 
language components (\eg lower layers learn language syntax or grammar while upper layers capture language semantics \cite{jawahar-etal-2019-bert}). We hypothesize that both in-file and cross-file context contribute to forming the understanding of language components. Therefore, we fuse the in-file and cross-file context at each Transformer layer so that generating the next token's hidden state will always depend on both context. At each time step $t$, for the $l$-th layer, we first compute the keys and values for cross-file and in-file context, using their $(l-1)$-th hidden states.

\vspace{-10pt} %
\begingroup\small\begin{gather*}
    K_\mathcal{C} = H^{[l-1]}_\mathcal{C}\mathbf{W}^K, V_\mathcal{C} = H^{[l-1]}_\mathcal{C}\mathbf{W}^V\\
    K_\mathcal{S}(t) = H_\mathcal{S}(t)^{[l-1]}\mathbf{W}^K, V_\mathcal{S}(t) = H_\mathcal{S}(t)^{[l-1]}\mathbf{W}^V
\end{gather*}\endgroup

Then, we concatenate the keys and values from both context so that, at time step $t$, the generating token can jointly attend them.

\vspace{-10pt} %
\begingroup\small\begin{gather*}
    K(t) = \texttt{concat}(K_\mathcal{C}, K_\mathcal{S}(t)), V(t) = \texttt{concat}(V_\mathcal{C}, V_\mathcal{S}(t))\\
    Q(t) = f_{\theta}(x_t)^{[l-1]}\mathbf{W}^Q, Attn(t) = \texttt{softmax}(\tfrac{Q(t)K(t)^\top}{\sqrt{d_K}})V(t)
\end{gather*}\endgroup

\section{Experiment Setup}
\label{sec:expr}

\subsection{Data}

Our data stem from the Python Package Index\footnote{\url{https://pypi.org/}}.
We collect permissively-licensed projects and filter out those with too few files ($\leq$5 python files) or too memory-consuming to build the project context graph ($\geq$5k nodes), ending up with 60,891 projects. Then, we divide the dataset into 80\%/10\%/10\% train, validation, and test sets. We notice that popular packages, such as \texttt{numpy}, are used as dependencies by many packages and will cause potential information leakage if \texttt{numpy} is part of the test set. Thus, we only include projects that were not used as dependencies by any training projects in the test set. We create prompts by cutting the source file at the location where completion requires cross-file context. See Appendix \ref{appendix:data-prep} for more details.

Figure~\ref{fig:motivation_example} shows an example prompt we create: it requires the details of \texttt{TagHandler} and \texttt{git} to complete the code accurately. In this work, we consider statement-level code completion, so the ground truth of the test sample is built accordingly. For the convenience of studying the model's prediction on local APIs (\ie APIs defined within the project), we further filter out the samples that either can not be parsed by the AST parser or do not include local API calls in the target statement (to be completed). Finally, we ended up with the 6,888 held-out prompts for evaluation.

\subsection{Implementation Details}
\paragraph{Cross-file Context} \ccfinder uses tree-sitter\footnote{\url{https://tree-sitter.github.io/}} to parse source code files. Tree-sitter is a widely used source code parser that generates the abstract syntax tree (AST) given a program. 
\ccfinder will traverse the AST to extract information as described in \S\ref{sec:context_construction}. Then, \ccfinder analyzes the \texttt{import} statements on top of import-dep\footnote{\url{https://pypi.org/project/import-deps/}}  to build the project context graph. In this work, we retrieve 2-hop neighbors with at max 128 project entities as cross-file context and each entity contains up to 128 tokens. These thresholds are data-driven 
to ensure the model input covers most of the relevant cross-file context (more details in Appendix~\ref{subsec:appendix_stats_retrieved_nodes}).

\paragraph{Model} The backbone of \tool is CodeGen \cite{nijkamp2022codegen} and we use CodeGen-350M-Mono for all experiments. In all settings, we finetune the model for 5 epochs with max sequence
length of 2,048 tokens and learning rate of 5e-5 with 5\% warm-up steps then cosine annealing.

\begin{table*}[!htp]\centering
\resizebox{0.85\textwidth}{!}{
\begin{tabular}{l c c c c c c c c}
\toprule
\multirow{2}{*}{Model} & \multirow{2}{*}{Finetuned} & \multirow{2}{*}{\shortstack{Cross-file\\ Entities}}
&\multicolumn{2}{c}{Code Match} &\multicolumn{3}{c}{ID Match} &\multirow{2}{*}{PPL ($\downarrow$)}\\
\cmidrule(lr){4-5}\cmidrule(lr){6-8}
&& &EM &BLEU-4 &EM &Prec. &Rec. \\\midrule
CodeGen &\myxmark & \myxmark &  14.56 & 33.12 & 22.91 & 47.74 & 50.75 & 2.88\\\hdashline\noalign{\vskip 0.4ex}
\quad + Finetune &\mycmark & \myxmark & 15.97 & 35.11 & 24.29 & 50.46 & 53.07 & 2.87\\
\quad \quad + Cross-file context &\mycmark & \mycmark & 17.00 & 36.34 & 25.80 & 48.91 & 54.76 & 2.77\\\midrule
\tool (Ours) &\mycmark & \mycmark & \textbf{21.39} & \textbf{41.65} & \textbf{31.26} & \textbf{55.45} & \textbf{57.83} & \textbf{2.69}\\
\bottomrule
\end{tabular}}
\caption{ 
Performance of \tool compared with baselines. We show that using the text prompt for cross-file entities (row 3) helps marginally compared to the in-file-only baseline (row 2). On the contrary, \tool with cross-file context (row 4) improves the performance by a large margin (+33.94\% Code Match EM and +28.69\% ID Match EM) compared to the in-file only baseline. In addition, we show that there is no degradation in perplexity (PPL) when evaluating all the tokens in the test set where the cross-file context is not always required, suggesting that adding cross-file context helps in general. See Appendix \ref{appendix: case_study} for additional case studies.
}\label{tab:main_result}
\vspace{-2pt}
\end{table*}

\subsection{Baselines \& Evaluation Metrics}
\label{subsec:baselines}
\paragraph{CodeGen} We consider two variations of the vanilla CodeGen model with in-file context only: (1) zero-shot, where we directly evaluate the pretrained CodeGen model on our test dataset, and (2) finetuned, where we finetune CodeGen on our dataset first and then evaluate. 

\paragraph{CodeGen w/ Cross-file Context} We also consider a prompting baseline where we prepend the cross-file context to the input sequence and finetune. Similar to the configuration of \tool, we reserve the first 128 tokens of the input for the code tokens from the cross-file context and use the rest tokens for the in-file context.

\paragraph{Evaluation Metrics}
We compute exact match (EM) and BLEU-4~\citep{kishore2002bleu} to assess the accuracy of the generated code.
While code match indicates the overall correctness of code completion, we want to zoom into the cases where cross-file context could most contribute, which is API usage. Therefore, we measure the identifier match to evaluate whether cross-file context improves the model's ability to predict the right APIs. To this end, we extract the identifiers from the model prediction and the ground truth, resulting in two ordered lists of identifiers. Then, we compare them and report the identifier match results in exact match, precision, and recall. 

Besides, we compute the perplexity of all the tokens on the test set to study whether adding cross-file context degrades performance when the cross-file context is not explicitly required.

\section{Results and Analysis}

\subsection{\tool Outperforms the Baselines}

We present the results in Table~\ref{tab:main_result}. \tool outperforms all baselines on all metrics with a clear margin, demonstrating the effectiveness of our proposed framework. We notice that when the cross-file context is prepended as a plain text prompt, CodeGen outperforms the other two baselines without cross-file context. However, limited by the maximum input length, it can only include a very limited amount of cross-file context, which significantly restricts its capacity. In contrast, \tool encodes the code sequence of an entity into one single token, enabling the model to incorporate more cross-file context while saving the input length. We present additional ablations for the baseline model in Appendix \ref{appendix:baseline}, and case studies in Appendix \ref{appendix: case_study}.

Besides, we see no degradation when the cross-file context is not explicitly required. We calculate the perplexity of all tokens in the test samples, regardless of whether they require cross-file context. We see that \tool achieves the lowest perplexity, indicating cross-file context in \tool is generally beneficial for code completion.

\subsection{\ccfinder Retrieves Relevant Cross-file Context}
\label{subsec:ccfinder_effectiveness}

The objective of \ccfinder is to locate and retrieve relevant
code context from other source files in the project. Identifiers (e.g., function names and parameters) are presumably one of the most critical API information. 
Therefore, we study the effectiveness of \ccfinder by assessing whether their retrieved context increases recall of the identifiers that appear in the ground truth.\footnote{We hypothesize that the inclusion of identifiers needed to complete a code is likely to benefit \tool.}
\begin{table}[!ht]
\centering
\resizebox{0.75\columnwidth}{!}{
\begin{tabular}{lc}\toprule
Code Context Type & ID Recall (\%) \\
\midrule
In-file context & %
75.19\\
In-file + Cross-file context & %
\textbf{95.55}\\
\bottomrule
\end{tabular}
}
\caption{\ccfinder retrieves 27.07\% more identifiers when compared to only in-file contexts.\vspace{-4pt}}
\label{tab:cov_gt_ids}
\end{table}

\begin{table}[!htbp]
\centering
\resizebox{0.9\columnwidth}{!}{
\begin{tabular}{l@{\hskip 0.1in} c@{\hskip 0.1in} c@{\hskip 0.1in} c@{\hskip 0.1in} c@{\hskip 0.1in} c}
\toprule
\multirow{2}{*}{Entities From}  &\multicolumn{2}{c}{Code Match} &\multicolumn{3}{c}{ID Match} \\\cmidrule(lr){2-3}\cmidrule(lr){4-6}
&EM &BLEU-4 &EM &Prec. &Rec. \\\midrule
Random & 15.68 & 35.23 & 24.07 & 49.75 & 52.69 \\
\ccfinder (1-hop) & 18.47 & 38.09 & 28.14 & 53.20 & 55.63 \\
\ccfinder (2-hop) & \textbf{21.39} & \textbf{41.65} & \textbf{31.26} & \textbf{55.45} & \textbf{57.83} \\
\bottomrule
\end{tabular}
}
\caption{Entities retrieved from \ccfinder are more useful than random entities, and 2-hop retrieval help achieve better performance.}
\vspace{-2pt}
\label{tab:random_entity}
\end{table}

Table~\ref{tab:cov_gt_ids} shows that the in-file context covers (recall) 75.19\% identifiers that appear in the ground truth.
In comparison, prompts augmented with retrieved cross-file identifiers bring up identifier recall to 95.55\%. This indicates that \ccfinder can retrieve most of the cross-file context that can help LM complete the input code. 
Note that while \ccfinder increases identifier recall by 27.07\%, Table \ref{tab:main_result} shows only a 7.08\% improvement in identifier recall. This indicates that building intelligent prompting techniques or training LMs to use cross-file context can lead to better performances. 
Further, Table~\ref{tab:random_entity} shows that random entities from the same project do not provide useful information since they are not necessarily related to the input code, and 2-hop retrieval outperforms 1-hop retrieval. These verify that \ccfinder retrieves relevant cross-file context and thus helps \tool. Appendix \ref{appendix:onehop} presents additional analysis.

\subsection{Entity Representation with \texttt{[SUM]} Token}
\label{subsec:sum}

We append a special token \texttt{[SUM]} to cross-file context to summarize their information (Figure~\ref{fig:model_arch}). Now, we study the importance of the \texttt{[SUM]} token for a better representation of cross-file context. As a comparison, we apply the widely-used mean pooling that takes the mean over every cross-file token's embedding as the cross-file representation. We train a \tool model with mean pooling and keep the rest of the settings the same. The result is in Table~\ref{tab:sum-rep}: our proposed \texttt{[SUM]} token effectively summarizes cross-file context and significantly outperforms the mean pooling strategy.

\begin{table}[!ht]
\centering
\resizebox{0.9\columnwidth}{!}{
\begin{tabular}{l@{\hskip 0.1in} c@{\hskip 0.1in} c@{\hskip 0.1in} c@{\hskip 0.1in} c@{\hskip 0.1in} c}
\toprule
\multirow{2}{*}{\tool}  &\multicolumn{2}{c}{Code Match} &\multicolumn{3}{c}{ID Match} \\\cmidrule(lr){2-3}\cmidrule(lr){4-6}
&EM &BLEU-4 &EM &Prec. &Rec. \\\midrule
Mean pooling & 16.78 & 36.02 & 25.01 & 50.50 & 52.61 \\
\texttt{[SUM]} & \textbf{21.39} & \textbf{41.65} & \textbf{31.26} & \textbf{55.45} & \textbf{57.83} \\
\bottomrule
\end{tabular}
}
\caption{Representing cross-file context with \texttt{[SUM]} token significantly outperforms mean pooling alternative.}
\label{tab:sum-rep}
\vspace{-2pt}
\end{table}

\subsection{Locales Help Learning Cross-file Context}
\label{subsec:impacts_cross_relation}

As introduced in \S\ref{subsec:model_input}, we prepend locales as relational hints for better entity representations. We study the effectiveness of such relational signals. As a comparison, we further study multi-task learning that encourages embedding relational information into entity representations.

\begin{table}[t]
\centering
\resizebox{0.9\columnwidth}{!}{
\begin{tabular}{l@{\hskip 0.1in} c@{\hskip 0.05in} c@{\hskip 0.05in} c@{\hskip 0.1in} c@{\hskip 0.05in} c}\toprule
\multirow{2}{*}{\tool}  &\multicolumn{2}{c}{Code Match} &\multicolumn{3}{c}{ID Match} \\
\cmidrule(lr){2-3}\cmidrule(lr){4-6}
&EM &BLEU-4 &EM &Prec. &Rec. \\\midrule
No Relations & 20.27 & 40.62 & 30.02 & 55.44 & 57.46 \\
MTL & 20.01 & 40.00 & 29.53 & 55.51 & 56.68\\
Locale &\textbf{21.39} & \textbf{41.65} & \textbf{31.26} & 55.45 & 57.83\\
Locale + MTL &21.25 & 41.44 & 31.05 & \textbf{55.83} & \textbf{58.03}\\
\bottomrule
\end{tabular}}
\caption{
Locales improve performance while learning cross-file relations with multi-task learning does not provide \tool more than marginal improvement. 
}
\vspace{-2mm}
\label{tab:cross_file_relation}
\end{table}

\paragraph{Multi-task w/ Edge Prediction} We use multi-task learning (MTL) to encode cross-file relations. Specifically, we train the model with an auxiliary edge prediction task among cross-file entities. We take representations of two cross-file entities generated by the LM layers and ask the model to predict what edge type connects them. 

\paragraph{Results} Table \ref{tab:cross_file_relation} presents the results.  While MTL achieves 97.2\% accuracy in the auxiliary edge prediction task, it hardly improves \tool in code completion. Such a gap suggests that even if MTL fulfills the expectation of embedding edge information, this information is not directly useful for code completion. 
In contrast, adding locales consistently improves \tool across all metrics. We hypothesize that this is due to locales providing an exact and direct signal as text (e.g., class\_name.method\_name). Thus the model could use them as \textit{short-cut} in code completion.

\section{Related Work}

In the last couple of years, a significant effort has been made to pretrain Transformer language models using unlabeled source code \cite{feng-etal-2020-codebert, ahmad-etal-2021-unified, wang-etal-2021-codet5, guo-etal-2022-unixcoder, ding-etal-2022-towards} to facilitate software engineering applications \cite{husain2019codesearchnet, iyer-etal-2018-mapping, tufano2019empirical, zhou2019devign}. Among these efforts, developing code generation models is noteworthy \cite{chen2021evaluating, xu2022systematic, gpt-j, black2021gpt, black2022gpt, nijkamp2022codegen, fried2022incoder, li2022competition}. 
Since most of these models are autoregressive language models, they can be directly used in code completion - given a code snippet as a prompt, generate the next $N$ tokens. 
Until recently, existing works in the literature use code snippet from the current file (where the user is writing code) to prompt the code generation models.
In a concurrent work, \citet{zhou2022doccoder} proposed to retrieve API documentation given a natural language (NL) intent and generate code based on them.
Our work has the same spirit as we propose to retrieve cross-file context (user-defined classes, functions from other project files) given a source code. The fundamental difference is that we utilize the \emph{import statements} for structured retrieval.

While the use of in-file or class context is rigorously studied for software engineering applications in the literature, the use of cross-file context is relatively under-explored in code completion backed by code LMs. Earlier works \cite{henninger1991retrieving, rosson1996reuse, michail2001codeweb, ye2000integrating, ye2002supporting, cubranic2003hipikat, inoue2003component, hill2004automatic, holmes2005using} in software engineering literature focused on developing tools to extract information from software repositories to help developers complete code fragments (e.g., variable, method name or body completion). 
On the other hand, recent works focus on modeling cross-file information in neural approaches. \citet{wang2021cocosum} proposed to model intra- and inter-class context for code summarization by extracting the Unified Modeling Language (UML) class diagrams.
A recent work \cite{shrivastava2022repository} proposed a prompt engineering technique that learns a repository-level prompt generator to generate example-specific prompts.
A concurrent work \cite{zhang2023repocoder} proposed an iterative retrieval-generation framework to augment prompt with cross-file context.

\vspace{-3pt}
\section{Conclusion}
The absence of \cfc for code language models (LMs) limits their practicality in modern software development.
In this work, we propose \tool, a framework that incorporates both in-file and cross-file context for code completion based on autoregressive code LMs. For this purpose, we build \ccfinder, a static code analysis tool that builds the project context graph, and find the most relevant cross-file context based on \texttt{import} statements. Empirical results show that \ccfinder successfully retrieves 27.07\% more relevant context that are not in the current file, and
our best \tool model achieves 33.94\% relative improvement over the in-file-context-only baseline.

\section*{Limitations}

\paragraph{Extension to other languages and third-party packages}
Our work focuses on Python language, which is widely used and has great availability of open-sourced software projects through PyPI. However, the main concept introduced in our work should be extensible to other languages. In addition, we focus on the project (repo) context in this work, and a potential extension is to incorporate third-party packages and building models to suggest the right third-party libraries to use. We leave these as future work.

\paragraph{Model performances with the absence of cross-file context}
In this work, we assumed that \tool could access the other source code files within the project to understand source code dependencies and utilize them accordingly to generate the target code completion. However, \tool may not access the code files in many cases, \eg users do not want an AI code LM to read their private or sensitive project APIs. Therefore, it is valid to ask -- how \tool performs when the cross-file context is absent. We evaluate \tool without access to cross-file context and compare with finetuned CodeGen model (second row in Table \ref{tab:main_result}). The results show that \tool performs 5--7\% lower (relative performance drop) than finetuned CodeGen model. Development of training strategies to bridge this performance gap is needed, and we leave this as future work.

\paragraph{Impact on different sized language models}
Although we use \texttt{CodeGen-350-mono} model in this work which consists of 350M parameters, we hypothesize that larger LMs (\eg 2B, 6B, or 16B variants of CodeGen) would result in similar or higher performance boost due to modeling cross-file context. However, we acknowledge that our work does not substantiate that our proposed technique would boost the performance of language models of any size.

\section*{Ethics Statement}
Our work aims at improving code generation with cross-file context to improve the usability of code LMs. We highlight the limitations of our work above. We do not expect our work to have a negative broader impact, though using code LMs always comes with certain risks, e.g., generating biased, toxic, and insecure code. We refer readers to Sec. 7 in \cite{chen2021evaluating} for a detailed discussion on the broader impact of code LMs. In addition, we reported our usage of computational resources in Appendix \ref{appendix:exp}.

\bibliography{main}
\bibliographystyle{acl_natbib}

\clearpage
\appendix

\section{Edge Types of Project Context Graph}
\label{subsec:appendix_edge_types}
We provide details of edge types that we use to build our project  context graph in Table~\ref{tab:edge_type}. The edges of the project context graph are directional, so for each edge type, we further define the expected entity type of its tail (\ie the entity that the edge is ``from") and head (\ie the entity that the edge is ``to") for each edge type. We also consider the reverse edges for certain types so that \ccfinder could retrieve entity siblings conveniently: \eg when a function is imported, \ccfinder could retrieve global variables it depends on by visiting the function's parent, \ie the file, with an edge of \texttt{Function Reverse} type, and reach the global variable with an edge of \texttt{Global Var.} type.

\begin{table}[h]
\centering
\resizebox{\columnwidth}{!}{
\begin{tabular}{lrr}\toprule
\textbf{Edge Type} & \textbf{Tail} & \textbf{Head} \\\midrule
Project File & root & file\\
Import & file & file \\
Global Var. & file & global var.\\
\makecell[l]{Global Var. Reverse} & global var. & file\\
Function & file & function \\
Function Reverse & function & file \\
Class & file & class \\
Class Reverse & class & file \\
Member Function & class & function\\
\bottomrule
\end{tabular}
}
\caption{The list of edge types we used to build the project context graph.}
\label{tab:edge_type}
\end{table}

\section{Statistics of Retrieved Entities}
\label{subsec:appendix_stats_retrieved_nodes}
This section presents the statistics of the retrieved entities of all samples in our dataset. Specifically, we hope to know two things that help us decide the experiment setup: (1) how many entities will be retrieved for the source file (2) how many tokens\footnote{Practically, these will be BPE sub-tokens from CodeGen tokenizer.} are there in the retrieved entity. Table~\ref{tab:entity_for_file} shows the ratio of source files that will retrieve project entities more than a specific threshold, and Table~\ref{tab:tokens_in_entity} ratio of entities that will contain tokens more than a specific threshold. \tool uses 128 as the maximum number of retrieved entities to be included in the model input and tokens within each entity. Consequently, \tool can always consider most of the cross-file context without causing too expensive computational overhead.

\begin{table}[h]
\centering
\resizebox{0.95\columnwidth}{!}{
\begin{tabular}{lrrrr}\toprule
Num of entities & > 32 & > 64 & > 128 & > 256\\\midrule
Ratio (\%) & 42.43 & 22.32 & 8.99 & 6.93\\
\bottomrule
\end{tabular}}
\caption{The ratio of the number of retrieved entities for the source code file with different thresholds.}
\label{tab:entity_for_file}
\end{table}

\begin{table}[ht]
\centering
\resizebox{0.95\columnwidth}{!}{
\begin{tabular}{lrrrr}\toprule
Num of tokens & > 32 & > 64 & > 128 & > 256\\\midrule
Ratio (\%) & 32.41 & 21.88 & 13.20 & 6.47\\
\bottomrule
\end{tabular}}
\caption{The ratio of the number of tokens within the retrieved entity with different thresholds.}
\label{tab:tokens_in_entity}
\end{table}

\section{Additional Details on Data Preprocessing}
\label{appendix:data-prep}
As we introduced in \S\ref{subsec:model_input}, the model input will be source code and its retrieved cross-file context. For training, we take the code files as samples. If a code file is too long, we split the code sequence into multiple chunks with the maximum length of the model input, and each chunk is paired with the same cross-file context. As the code files are from distinct PyPI projects, we assume the duplicated samples should be rare.

We follow the standard code completion setting for testing~\cite{nijkamp2022codegen, xu2022systematic}, creating incomplete programs as prompts and asking the model to predict the following pieces of code. Specifically, we create prompts by cutting the source file at the location where completion requires cross-file context. We present the details of samples' sequence length, in terms of BPE sub-tokens, in Table~\ref{tab:prompts_stats}. For cross-file context, we concatenate the text of all retrieved entities as a sequence and count the length. We could see that the cross-file context is typically long and could not be consumed as plain text together with the prompts. \tool designs to compress the entities with the special token \texttt{[SUM]} that effectively alleviates this limitation.

\begin{table}[h]
\centering
\resizebox{\columnwidth}{!}{
\begin{tabular}{lrrrr}\toprule
 & Mean & Max & Median & Min\\\midrule
Prompts & 1,354 & 32,599 & 758 & 7\\
Cross-file context & 4,485 & 186,339 & 1,928 & 22\\
\bottomrule
\end{tabular}}
\caption{Length statistics of prompts and cross-file context of the test set.}
\label{tab:prompts_stats}
\end{table}

\section{Additional Details on Experiments}
\label{appendix:exp}
Our code is based on Transformers \cite{wolf-etal-2020-transformers}. We train our models on a machine with 8 Nvidia A100s. Each job takes around 50 hours (i.e., 400 GPU hours) to train. We perform one round of experiments only as it is very expensive to repeat the experiments many times. The hyperparameter used is from our initial small-scale grid search on hyperparameters, where we find that the final performance is relatively stable. 

\section{Additional Ablation Studies}
\label{appendix:ablation}
\subsection{$k$-hop Retrieval}
\label{appendix:onehop}

As we see from Table \ref{tab:random_entity}, $k=1$ underperforms comparing to $k=2$. This is because $k=1$ fetches less comprehensive context. For example, with the import statement \texttt{import FileA as A}, we can access class \texttt{X}’s static member function \texttt{Y} as: \texttt{A.classX.funcY} through 2-hop retrieval, whereas 1-hop retrieval will not fetch. In fact, 1-hop retrieval won’t fetch any class member function if only the file is imported, which happens frequently in Python. Given the great coverage of $k=2$ (Table \ref{tab:cov_gt_ids}) and given we found too many unrelated entities were retrieved if we use $k>2$, we decided to use $k=2$ throughout the work. 

\subsection{Additional Baseline Variants}
\label{appendix:baseline}
In addition to the CodeGen w/ Cross-file Context baseline (\S\ref{subsec:baselines}) which uses the same cross-file context tokens as in \tool, we experimented with a simplified setting that only takes the locales and the signature prototypes (name, arguments, and default return types, if present) to fit in more cross-file context within the limited token space. 

\begin{table}[!ht]
\centering
\resizebox{\columnwidth}{!}{
\begin{tabular}{l@{\hskip 0.1in} c@{\hskip 0.1in} c@{\hskip 0.1in} c@{\hskip 0.1in} c@{\hskip 0.1in} c}
\toprule
\multirow{2}{*}{}  &\multicolumn{2}{c}{Code Match} &\multicolumn{3}{c}{ID Match} \\\cmidrule(lr){2-3}\cmidrule(lr){4-6}
&EM &BLEU-4 &EM &Prec. &Rec. \\\midrule
CodeGen + Finetune &&&&& \\
\quad + Cross-file context (full) & 17.00 & 36.34 & 25.80 & 48.91 & 54.76 \\
\quad + Cross-file context (simp.) & 17.49 & 37.57 & 26.76 & 51.71 & 54.88 \\
\tool & \textbf{21.39} & \textbf{41.65} & \textbf{31.26} & \textbf{55.45} & \textbf{57.83} \\
\bottomrule
\end{tabular}
}
\caption{All baselines significantly outperforms \tool.}
\vspace{-2pt}
\label{tab:additional_baseline}
\end{table}

We see the performance only improves marginally when using simplified cross-file context, and it still underperforms \tool significantly. This suggests that baseline models have substantial limitations of sequence lengths that the performance is subpar even if we simplify the cross-file context, while \tool is capable of compressing up to 16,384 (=128x128) tokens of cross-file context into only 128 vectors, making cross-file context readily available for the model to use. 

\section{Case Study}
\label{appendix: case_study}
We present a case study by comparing \tool with the finetuned CodeGen model with and without cross-file context. 

\subsection{\tool vs. CodeGen finetuned w/o cross-file context}
We present two qualitative examples in Figure \ref{fig:qual_ex1} and \ref{fig:qual_ex2} where Baseline refers to the CodeGen model finetuned w/o cross-file context.
From Figure \ref{fig:qual_ex1}, we see that Baseline calls \texttt{get\_credentials()} function of \texttt{AssumeRoleExecutor} class, which does not exist. On the other hand, being able to access cross-file context, \tool predicts the correct function \texttt{execute()} that returns \texttt{Credentials} class instance. In Figure \ref{fig:qual_ex2}, we see similar behavior from Baseline as it predicts \texttt{get\_key\_and\_secret} method. In language generation literature, such an inaccurate or unfaithful generation given the input is known as \emph{hallucination} \cite{ji2022survey}.

\subsection{\tool vs. CodeGen finetuned with cross-file context}
We present two qualitative examples in Figure \ref{fig:qual_ex3} and \ref{fig:qual_ex4} where Baseline refers to the CodeGen model finetuned with cross-file context.
From Figure \ref{fig:qual_ex3}, we see that Baseline calls \texttt{parse()} method of \texttt{CommandLineArgs} class, which does not exist. On the other hand, due to providing cross-file context, \tool predicts the correct function \texttt{get\_cli\_args()} that returns \texttt{CliArgs} class instance. In Figure \ref{fig:qual_ex4}, we see similar behavior from Baseline as it predicts \texttt{update\_from\_dict(schedule)} method. In both cases, the Baseline fails to make an accurate prediction due to the truncation of cross-file context, while \tool predicts correctly as it encodes each cross-file context entity individually and then utilizes their embedding in the self-attention mechanism. In contrast to cross-file context truncation, we could improve the Baseline by effectively selecting the most useful entities from the ordered list of retrieved cross-file entities. We leave this as our future work.

\clearpage

\begin{figure*}[t]
    \centering
    \includegraphics[width=0.85\textwidth]{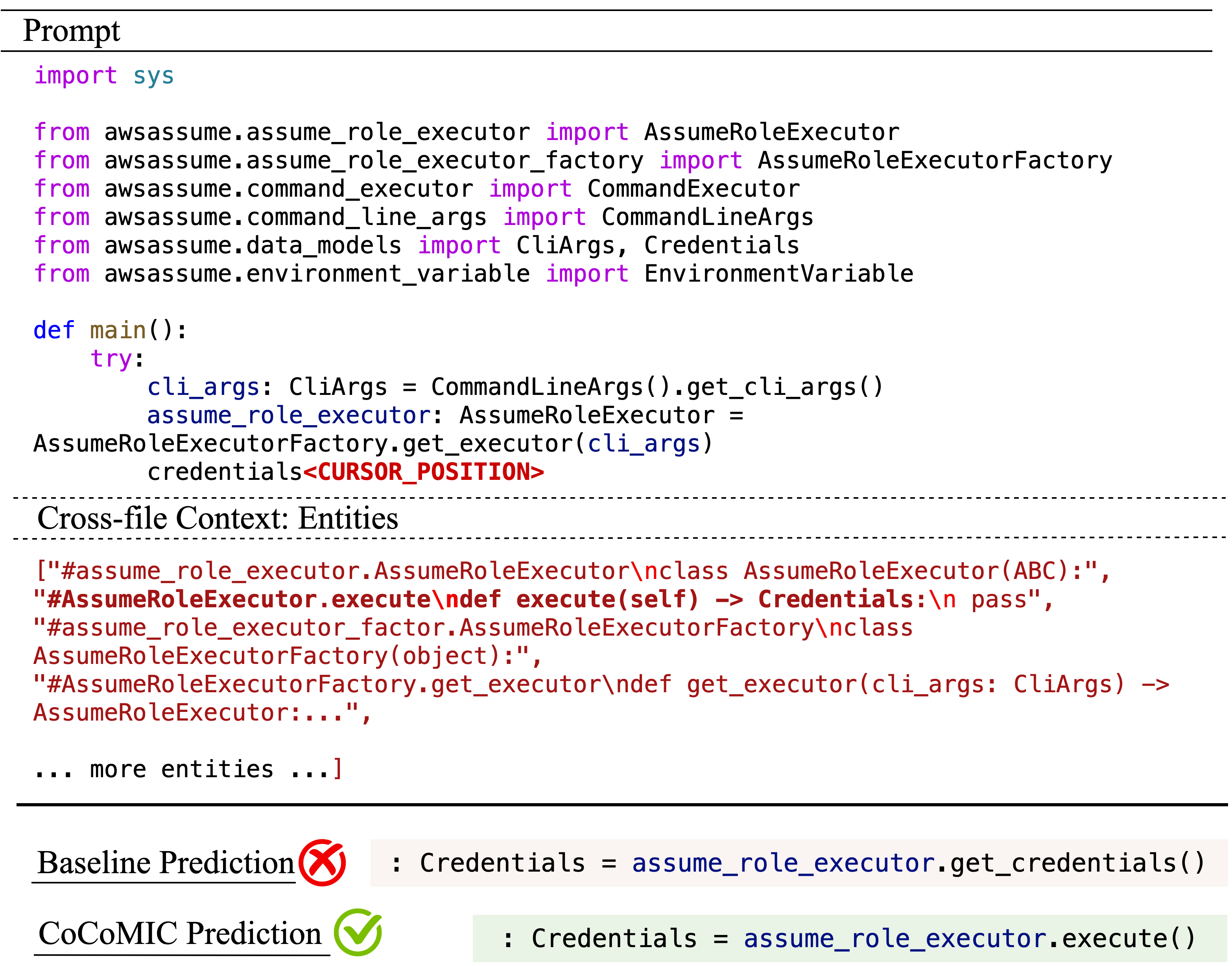}
    \caption{\tool \emph{vs.} CodeGen finetuned \textbf{w/o} cross-file context: qualitative example-1.}
    \label{fig:qual_ex1}
\end{figure*}

\begin{figure*}[ht]
    \centering
    \includegraphics[width=0.85\textwidth]{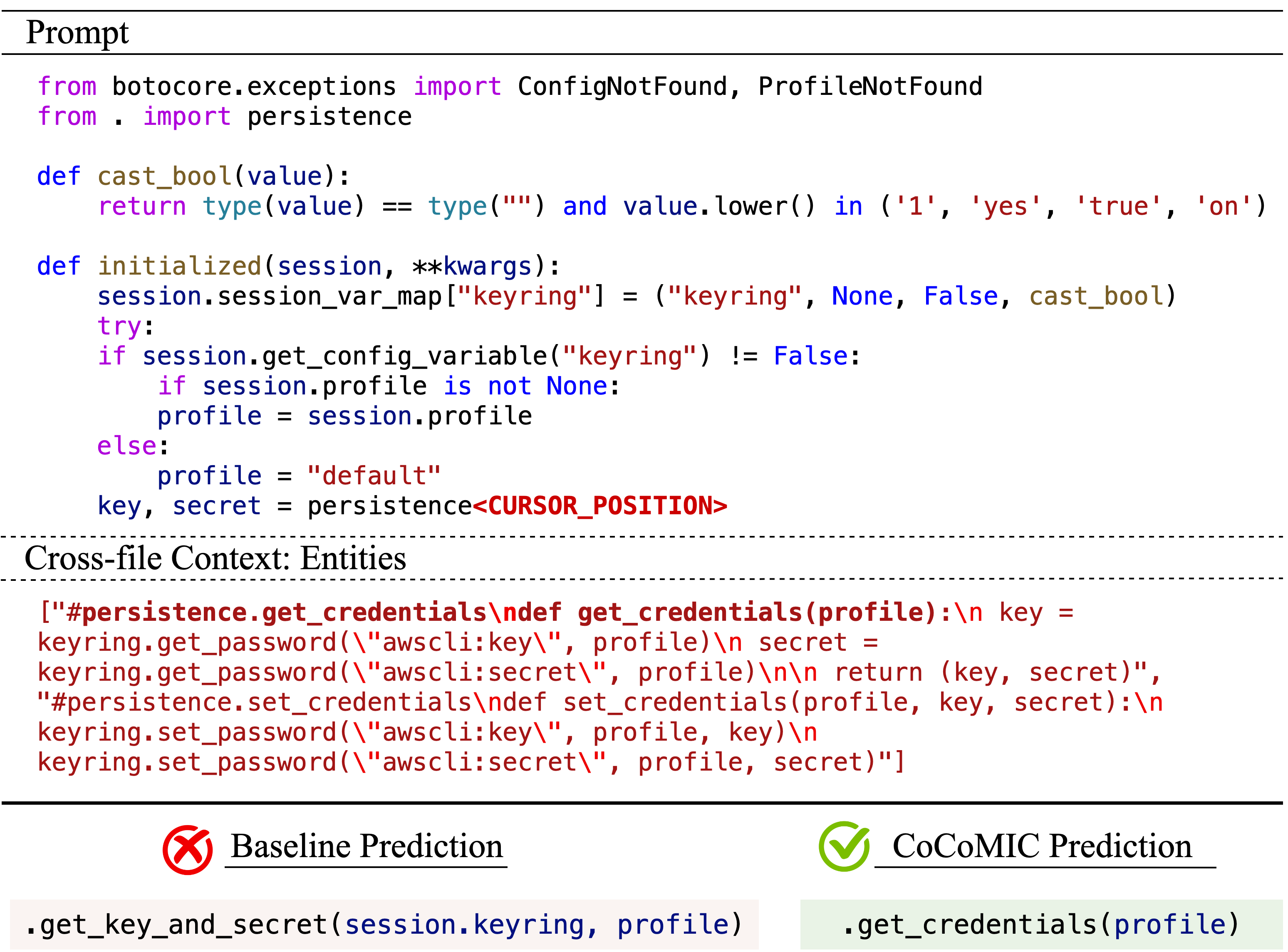}
    \caption{\tool \emph{vs.} CodeGen finetuned \textbf{w/o} cross-file context: qualitative example-2.}
    \label{fig:qual_ex2}
\end{figure*}

\begin{figure*}[t]
    \centering
    \includegraphics[width=0.9\textwidth]{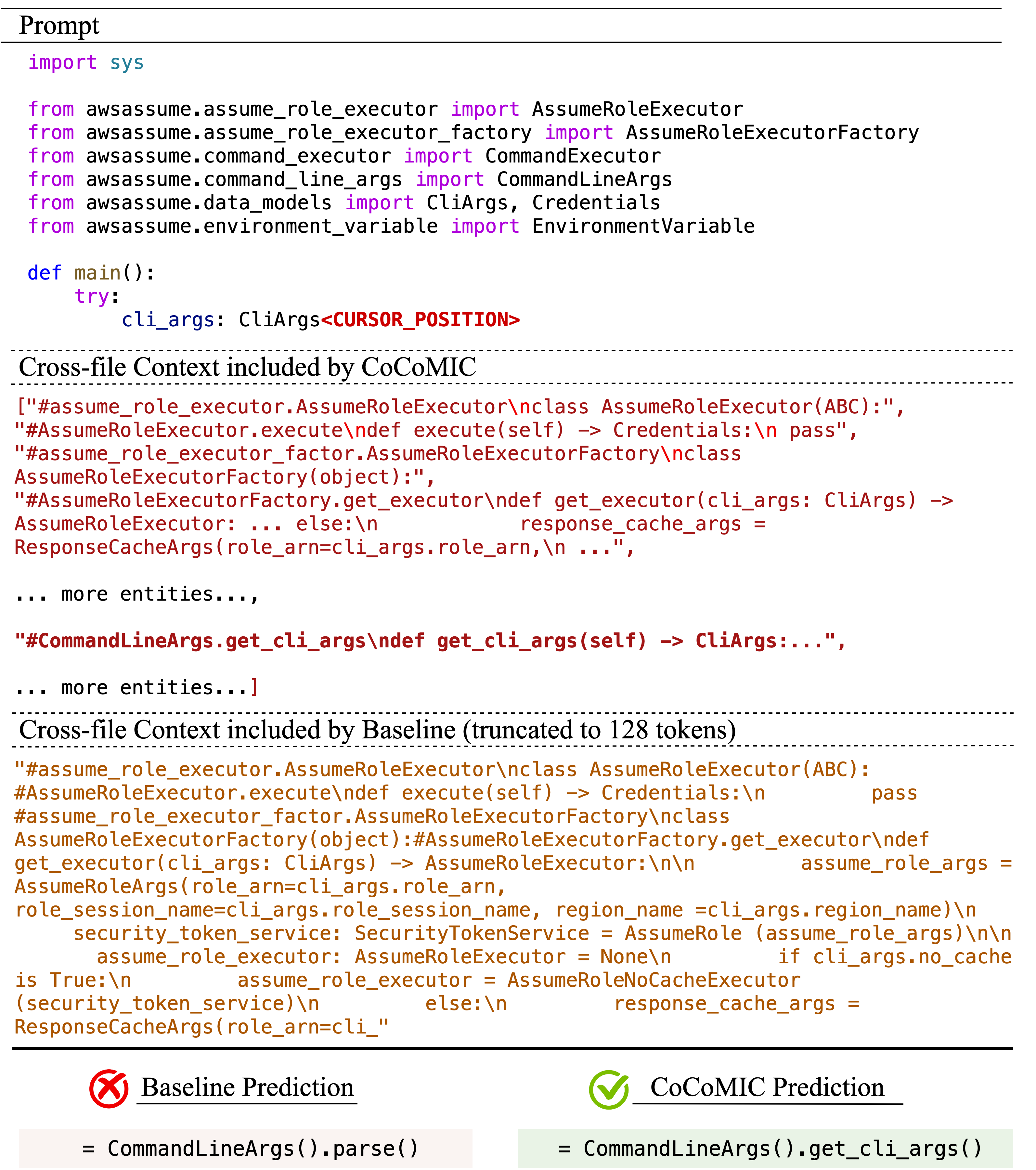}
    \caption{\tool \emph{vs.} CodeGen finetuned \textbf{with} cross-file context: qualitative example-1.}
    \label{fig:qual_ex3}
\end{figure*}

\begin{figure*}[t]
    \centering
    \includegraphics[width=0.9\textwidth]{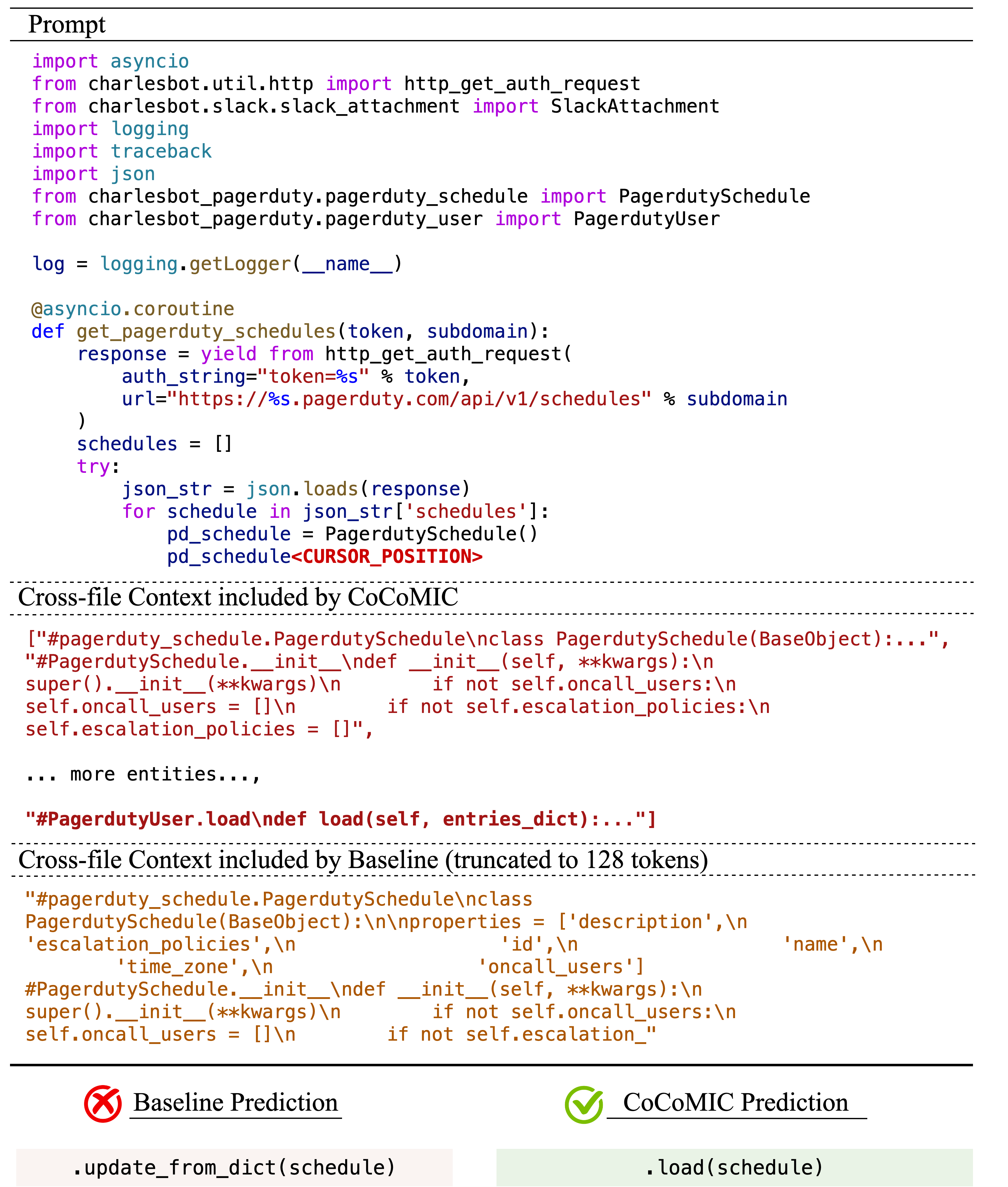}
    \caption{\tool \emph{vs.} CodeGen finetuned \textbf{with} cross-file context: qualitative example-2.}
    \label{fig:qual_ex4}
\end{figure*}

\end{document}